\pdfoutput=1

\documentclass[11pt]{article}

\usepackage{acl}

\usepackage{times}
\usepackage{latexsym}

\usepackage[T1]{fontenc}

\usepackage[utf8]{inputenc}

\usepackage{microtype}

\usepackage{amsmath} 
\usepackage{amsthm} 
\usepackage{amsfonts} 
\usepackage{amssymb}
\usepackage{multirow}
\usepackage{dcolumn}
\usepackage{array}
\usepackage{graphicx}

\usepackage{algorithm}
\usepackage[noend]{algpseudocode}

\usepackage{graphicx} 
\usepackage{comment}
\usepackage[inline]{enumitem}
\usepackage{flushend}
\usepackage{amsmath}
\usepackage{amssymb}
\usepackage{colortbl}
\usepackage{cellspace}
\usepackage{xcolor}
\usepackage{booktabs}

\usepackage{enumitem}
\usepackage{url}
\usepackage{graphicx}
\usepackage{caption}
\usepackage{booktabs}

\title{A Comprehensive Study of Gender Bias in Chemical Named Entity Recognition Models}

\author{Xingmeng Zhao, Ali Niazi, and Anthony Rios \\
    Department of Information Systems and Cyber Security\\
  The University of Texas at San Antonio\\
  \texttt{\{xingmeng.zhao, ali.niazi, anthony.rios\}@utsa.edu} \\}

\begin{document}
\maketitle

\begin{abstract}

Chemical named entity recognition (NER) models are used in many downstream tasks, from adverse drug reaction identification to pharmacoepidemiology. However, it is unknown whether these models work the same for everyone. Performance disparities can potentially cause harm rather than the intended good. This paper assesses gender-related performance disparities in chemical NER systems. We develop a framework for measuring gender bias in chemical NER models using synthetic data and a newly annotated corpus of over 92,405 words with self-identified gender information from Reddit. Our evaluation of multiple biomedical NER models reveals evident biases. For instance, synthetic data suggests female-related names are frequently misclassified as chemicals, especially for brand name mentions. Additionally, we observe performance disparities between female- and male-associated data in both datasets. Many systems fail to detect contraceptives such as birth control. Our findings emphasize the biases in chemical NER models, urging practitioners to account for these biases in downstream applications.
\end{abstract}

\section{INTRODUCTION}

Chemical named entity recognition (NER) is the extraction of chemical mentions (e.g., drug names) from the text. Chemical NER is essential in many downstream tasks, from pharmacovigilance~\cite{2014pharmacovigilance} to facilitating drug discovery by mining biomedical research articles~\cite{Agarwal2008LiteratureMI}. For instance, Chemical NER systems are the first step in pipelines developed to mine adverse drug reactions (ADRs)~\cite{Farrugia2020MiningDI,Mamm2013PharmacovigilanceIP}.
However, it is unknown whether these systems perform the same for everyone. Who benefits from these systems, and who can be harmed? In this paper, we present a comprehensive analysis of gender-related performance disparities of Chemical NER Systems.


Performance disparities have recently received substantial attention in the field of NLP. For example, there are differences in text classification models across sub-populations such as gender, race, and minority dialects~\cite{dixon2018measuring, park2018reducing, badjatiya2019stereotypical, rios2020fuzze, lwowski2021risk, mozafari2020hate}. Performance disparities can manifest in multiple parts of NLP systems, including the pre-trained models (e.g., word embeddings) and their downstream applications~\cite{zhao-etal-2019-gender,goldfarb2021intrinsic,zhao2017men}. While previous research has explored these disparities for NER systems, the focus has been largely on synthetic data and non-biomedical NER applications~\cite{Mehrabi2020ManIT}. Our study addresses this gap by providing a comprehensive examination of gender-related performance disparities in Chemical NER, focusing on both synthetic and real-world data.

This paper is most similar to \citet{Mehrabi2020ManIT} with two primary distinctions. First, our focus is on Chemical NER, a less studied area in Biomedical NLP despite its having major bias implications. Second, while \citet{Mehrabi2020ManIT} uses synthetic data and templates (e.g., NAME in LOCATION) for bias analysis, we delve deeper into the potential including an analysis of the interaction of morphology patterns on bias. For instance, \citet{lieven2015effect} highlighted a preference for linguistically feminine brand names in the market, leading drug companies to adopt such naming conventions. These patterns in training data can inadvertently cause models to misclassify female names as chemicals.

We also examine real-world data looking at the performance of chemical NER systems on groups that identify as male or female. For instance, \citet{sundbom2017men} shows that women are more frequently prescribed antidepressants than men. Other studies, like \citet{riley1998sex}, reveal gender differences in pain sensitivity and opioid prescriptions, with women receiving opioids twice as often. If chemical NER models struggle to detect the drugs often mentioned, then it may cause gender-specific biases in their performance. Our analysis identifies some of these patterns in real data. 

Overall, this paper presents a dual approach: we explore template data but also assemble and annotate a novel real-world dataset with self-identified gender information. \footnote{The dataset and code will be released publicly upon acceptance.} Synthetic data allows us to target specific biases in the models (e.g., morphological issues).
Likewise, we believe exploring data from people who have self-identified their demographic information will provide a more realistic understanding of how these models will perform based on how people write and what they write about. 

Our main contributions are two-fold:
\vspace{-.2cm}
\begin{enumerate}
\item We introduce a novel annotated Chemical NER dataset for social media data. Moreover, the dataset contains self-identified gender information to be used to measure gender bias in Chemical NER models. To the best of our knowledge, this is the first Reddit-based Chemical NER dataset. \textit{Moreover, it is the first Chemical NER dataset with self-identified gender information.}
\vspace{-.2cm}
\item We provide a comprehensive testing framework for gender bias in Chemical NER using both synthetic and real-world data. To the best of our knowledge, our results are the first to conduct bias analysis for chemical NER models. This allows a better understanding of modern chemical NER techniques.

\end{enumerate}

\section{RELATED WORK}

Prior work extensively curated labeled data for chemical NER and developed domain-specific models. For example, the CHEMDNER corpus~\cite{krallinger2015chemdner} was created for the 2014 BioCreative shared task on chemical extraction from text. Researchers recognize the importance of these systems and are working to make them as fair and accurate as possible. Likewise, the CDR~\cite{li2016biocreative} dataset was developed to detect chemical-disease relations for the 2015 shared task. Similar to traditional NER tasks~\cite{li2020survey}, a broad range of approaches have been proposed to detect Chemicals~\cite{rocktaschel2012chemspot,chiu2021recognizing, lee2020biobert,sun2021deep,lopez2021combining,weber2021hunflair}, from traditional conditional random fields to deep learning methods. Many recent neural network-based advances can be broken into three main groups of models, word, character, and contextual embedding-based models. For instance, \citet{lee2020biobert} trained a biomedical-specific BERT model that improved on many prior state-of-the-art results. HunFlair~\cite{weber2021hunflair} introduced a method that matches the word, contextual, and character embeddings into a unified framework to achieve state-of-the-art performance. In this paper, we evaluate several state-of-the-art systems. Particularly, we focus on systems that use word embeddings, sub-word embeddings, and character embeddings, which allows us to understand the impact of morphological features of the chemical names on gender bias.

Several previous works have measured and highlighted bias in different NLP tasks. For instance, ~\citet{sap2019risk} measures the bias of offensive language detection models on African American English. Likewise, \citet{park2018reducing} measures gender bias of abusive language detection models and evaluates various methods such as word embedding debiasing and data augmentation to improve biased methods. \citet{davidson2019racial} shows racial and ethnic bias when identifying hate speech online and that tweets in the black-aligned corpus are more likely to be assigned hate speech. \citet{gaut2020towards} creates the WikiGenderBias dataset to evaluate the gender bias in the relation extraction (RE) model, confirming that the RE system behaves differently when the target entities are of different genders. \citet{cirillo2020sex} demonstrate that biases in biomedical applications can stem from various sources, such as skewed diagnoses resulting from clinical depression scales that measure symptoms more prevalent in women, potentially leading to a higher reported incidence of depression among this group~\cite{martin2013experience}. Other sources include the underrepresentation of minority populations such as pregnant women~\cite{world2009mental}, non-representative samples in AI training data, and inherent algorithmic discrimination, all potentially contributing to inaccurate and unfair results.

Overall, several metrics have been proposed to measure gender bias. One of the most commonly used metrics involves measuring bias by examining model performance disparities on male and female data points~\cite{kiritchenko2018examining}. Performance disparities have been observed across a wide array of NLP tasks such as detecting virus-related text~\cite{lwowski2021risk}, language generation~\cite{sheng2019woman}, coreference resolution~\cite{zhao2018gender}, named entity recognition~\cite{Mehrabi2020ManIT}, and machine translation~\cite{font2019equalizing}. Most related to this study, researchers have shown that traditional NER systems (i.e., to detect people, locations, and organizations) are biased concerning gender~\cite{Mehrabi2020ManIT}. Specifically, \citet{Mehrabi2020ManIT} demonstrates that female-related names are more likely to be misidentified as a location than male names. This stream of research underscores the importance of our investigation into performance disparities in NLP.

Finally, while not directly studied in prior NER experiments. It is important to discuss some background about morphological elements of chemical names. Morphological elements often representing masculinity or femininity are frequently used in chemical naming conventions. According to \citet{lieven2015effect}, consumers perceive linguistically feminine brand names as warmer and likelier. For instance, adding a diminutive suffix to the masculine form of the name usually feminizes it. The masculine names such as Robert, Julius, Antonio, and Carolus (more commonly Charles today) are feminized by adding the suffixes ``a'', ``ia'', ``ina'', or ``ine'' to generate Julia, Roberta, Antonia, and Caroline, respectively. The suffixes ``ia'' and ``a'' is commonly used for inorganic oxides such as magnesia, zirconia, silica, and titania~\cite{hepler2015just}. Likewise, ``ine'' is used as the suffix in many organic bases and base substances such as quinine, morphine, guanidine, xanthine, pyrimidine, and pyridine. Hence, while these practices were not originally ``biased'' in their original usage, they can potentially impact model performance (e.g., feminine names can be detected as chemicals). Therefore, the patterns can cause biased models. As part of our approach to investigate this potential source of bias, we propose using synthetic data to quantify this phenomenon. 

\begin{table}[t]
\centering
\resizebox{\linewidth}{!}{
\begin{tabular}{lrrr}
\toprule
                 &\textbf{\# of Chemical Mentions} & \textbf{\# Sentences} & \textbf{\# Words}   \\ \midrule
\textbf{CDR}       &   4,409    &  14,306   &  346,001 \\
\textbf{CHEMDNER}  &   84,355   &  87,125  &   2,431,247 \\
\textbf{CHEBI} &  24,121 &  12,913  &   423,577  \\ \midrule

\textbf{AskDoc MALE} &    1,501   &  2,862   & 52,221  \\ 
\textbf{AskDoc FEMALE} &   1,774    & 2,151   & 40,184 \\ 
\textbf{AskDoc ALL} &  3,275     &  5,013  &  92,405 \\ \midrule
\textbf{Synthetic MALE} &   2,800,000    &  2,800,000  & 25,760,000  \\
\textbf{Synthetic FEMALE} &   2,800,000   &  2,800,000  &  25,760,000 \\
\textbf{Synthetic ALL} &    5,600,000   & 5,600,000   &  51,520,000 \\ \bottomrule
\end{tabular}} 
\caption{Dataset statistics.\vspace{-2em}}
\label{tab:stats}
\end{table}

\section{DATASETS}

We use five main datasets used in our experiments: three are publicly-released datasets based on PubMed (CDR~\cite{li2016biocreative}, CHEMDNER~\cite{krallinger2015chemdner}, and CHEBI~\cite{shardlow2018new}) and two are newly curated datasets, one using social media data and another based on templates. Table~\ref{tab:stats} provides their statistics. We selected the PubMed datasets for their prominence in chemical NER research. At the same time, the r/AskDocs subreddit was chosen for its large community, diverse health discussions, and consistent gender identification format, such as ``I [25 M]''. We provide complete descriptions of the publicly-released datasets in the Appendix. In this section, focus on the description of the newly collected and annotated data.



\paragraph{Synthetic (Template) Data} We designed a new synthetic dataset to quantify the gender bias in the Chemical NER models. Intuitively, the purpose of the synthetic dataset is to measure two items. First, do gender-related names and pronouns get incorrectly classified as Chemicals (i.e., cause false positives)? Second, does the appearance of gender-related names/pronouns impact the prediction of other words (i.e., cause false negatives)? Specifically, we create templates such as ``[NAME] said they have been taking [CHEMICAL] for an illness.'' In the ``[NAME]'' column, we filled in the names associated with the male and female genders based on the 200 most popular baby names provided by the Social Security Administration~\footnote{\url{https://www.ssa.gov/oact/babynames/}}. Hence, we refer to these ``gender-related'' names in this paper. We recognize that gender is not binary and that names do not equal gender. We also recognize that the names do not accurately capture immigrants.  This is a similar framework used by \citet{mishra2020assessing} and other gender bias papers~\cite{kiritchenko2018examining}. The ``[CHEMICAL]'' field is filled with the chemicals listed in the Unified Medical Language System (UMLS)~\cite{bodenreider2004unified}. For example, completed templates include  ``John said they has been taking citalopram for illness.'' and ``Karen said they has been taking citalopram for illness.'' We created examples using five templates, 200 chemicals, and 200 names for each gender for each decade from 1880 to 2010, generating a total of 200,000 templates for each of the 14 decades. A list of additional templates is shown in Table~\ref{tab:templates}. This dataset is only used for evaluation.

\begin{table}[t]
\centering
\resizebox{\columnwidth}{!}{%
\renewcommand{\arraystretch}{1.25}
\begin{tabular}{l}
\toprule
\textbf{Templates} \\ \midrule
{[}NAME{]} said they have been taking {[}CHEMICAL{]} for an illness.     \\ 
Did you hear that {[}NAME{]} has been using {[}CHEMICAL{]}.          \\ 
{[}CHEMICAL{]} has really been harming {[}NAME{]}, I hope they stop. \\ 
I think {[}NAME{]} is addicted to {[}CHEMICAL{]}. \\ 
{[}NAME{]}, please stop taking {[}CHEMICAL{]}, it is bad for you.  \\  \bottomrule
\end{tabular}}
\caption{Templates used to create the synthetic dataset.\vspace{-1em}}
\label{tab:templates}
\end{table}

\paragraph{AskDocs} We develop a new corpus using data from the Reddit community r/AskDocs. r/AskDocs provides a platform for peer-to-peer and patient-provider interactions on social media to ask medical-related questions. The providers are generally verified medical professionals. We collected all the posts from the community with self-identified gender mentions. To identify self-identified gender, we use a simple regular expression that looks for mentions of ``I'' or ``My'' followed by gender, and optionally age, e.g., ``I [F34]'', ``My (23F)'', ``I [M]''. Next, following general annotation recommendations for NLP~\cite{pustejovsky2012natural}, the annotation process was completed in two stages to increase the reliability of the labels. First, two graduate students annotated chemicals in the dataset resulting in an inter-annotator agreement of .874, achieving a similar agreement score as CDR and CHEMDNER. Second, a graduate student manually reviewed all disagreeing items to adjudicate the label and generate the gold standard. All students followed the same annotation guidelines developed for the CHEMDNER corpus. Contrary to the synthetic dataset, the actual data will allow users to measure biases arising from text content differences across posts with different self-identified gender mentions.


\section{Methods} 


The goal of NER is to classify words into a sequence of labels. Formally, given an input sequence $\mathcal{X} = [x_1, x_2, \ldots, x_N]$ with N tokens, the goal of NER is to output the corresponding label sequence $\mathcal{Y} =[y_1, y_2, \ldots, y_N]$ with the same length, thus modeling the probabilities over a sequence $p(\mathcal{Y} | \mathcal{X} )$.
For this task, we conducted an experiment evaluating out-of-domain models on the AskDoc corpus. Specifically, models were trained and optimized on the CHEMDNER and CDR datasets and then applied to the AskDoc dataset.
All models are evaluated using precision, recall, and F1. To measure bias, we use precision, recall, and F1 differences~\cite{czarnowska2021quantifying}. Specifically, let $m$ be Males' performance metric (e.g., F1), and $f$ represent the Female metric. The bias is measured using the difference $f-m$.

\subsection{MODELS}

We evaluate three distinct models: Word Embedding models~\cite{mikolov2013distributed}, Flair embedding models~\cite{akbik2018coling}, and BERT-based models~\cite{devlin2019bert}. While the embeddings for each model type vary, the sequence processing component is the same for each method. Specifically, following best practices for state-of-the-art NER models~\cite{akbik2019flair}, we use a Bidirectional long short-term memory network (Bi-LSTM)~\cite{hochreiter1997long} due to its sequential characteristics and capability to capture long-term dependencies. Recent research has shown that Bi-LSTM models can produce state-of-the-art performance when combined with contextual embeddings and Conditional Random Fields (CRFs)~\cite{mueller2020sources,veyseh2022generating}. Hence, in this paper, we use the Bi-LSTM+CRF implementation in the Flair NLP framework~\cite{akbik2019flair}. The Bi-LSTM+CRF model is flexible because it can accept arbitrary embeddings as input. It is not constrained to traditional word embeddings (e.g., Word2Vec). We describe the embeddings we experiment with in the next Section.

\subsection{EMBEDDINGS}

We explore three sets of embeddings: Word2Vec, Flair, and BERT. For all embeddings, we experiment with domain-specific (e.g., trained on PubMed) and general embeddings (e.g., Google News corpus). We chose these three embedding types because they cover word, subword, and character-level embedding methods. Social media texts are brief and informal. Drugs and chemicals are typically described in descriptive, nontechnical language with spelling errors. These issues challenge social media Chemical NER. Moreover, some medications, like ``all-trans-retinoic acid'', contain morphologically difficult parts. Yet, similar-structured phrases still generally represent similar things~\cite{zhang2021identifying}. How we represent words can directly impact performance and bias. We describe each embedding we use below:

\paragraph{Word2Vec.} We use Word2Vec domain-specific embeddings pre-trained on PubMed and PubMed Central~\cite{pyysalo2013distributional} and general embeddings trained on the Google News corpus~\cite{wordvecgoog}. The embeddings are publicly released as part of the FLAIR package. It is important to state that word embeddings have a major limitation. Word embeddings use a distinct vector to represent each word and ignore words' internal structure (morphology). This can result in models not particularly good at learning rare or out-of-vocabulary (OOV) words in the data. The growing number of emerging chemicals/drugs with diverse morphological forms makes recognizing chemical entities on social media platforms particularly challenging. Another challenge posed by user-generated content is its unique characteristics and use of informal language, typically short context, noisy, sparse, and ambiguous content. Hence, we hypothesize that word embeddings would perform worse than other methods. However, it is unclear how these differences can impact bias.

\paragraph{Flair/HunFlair.} \citet{weber2021hunflair} and \citet{akbik2019pooled} recently proposed a Flair contextual string embeddings (a character-level language model). Specifically, we use two versions of the embeddings in the HunFlair extension of the Flair package~\cite{weber2021hunflair}. The domain-specific embeddings are pre-trained on a corpus of three million full-text articles from the Pubmed Central BioC text mining collection~\cite{comeau2019pmc} and about twenty-five million abstracts from PubMed. The general embeddings are trained on a one billion word news corpus~\cite{akbik2019pooled}.

Unlike word embeddings mentioned above, Flair embeddings are a contextualized character-level representation. Flair embeddings are obtained from the hidden states of a bi-directional recurrent neural network (BiRNN). They are trained without any explicit notion of a word. Instead, Flair models a word as sequences of characters. Moreover, these embeddings are determined by the text surrounding them, i.e., the same word will have different embeddings depending on its contextual usage. The variant of the Flair embedding used in this study is the Pooled Flair embedding~\cite{weber2021hunflair,akbik2018coling}. Furthermore, we use the forward and backward representations of Flair embeddings returned from the BiRNN. Intuitively, character-level embeddings can potentially help improve model predictions with better OOV handling.

\paragraph{(Bio)BERT.} We also evaluate two transformer-based embeddings: BERT and BioBERT. Specifically, we use the BERT variant ``bert-base-uncased'' available Flair and HuggingFace~\cite{wolf-etal-2020-transformers}. BERT was pre-trained using the BooksCorpus (800M words) and English Wikipedia (2,500M words)~\cite{devlin-etal-2019-bert}. Likewise, BioBERT embeddings further fine-tuned BERT on PubMed~\cite{lee2020biobert}.

BERT embeddings are based on subword tokenization, so BERT can potentially handle OOV better than word embeddings alone. Intuitively, it fits somewhere between Flair (generating word embeddings from character representations) and Word2Vec (which independently learns embeddings for each word). Likewise, each word representation is context-dependent. Hence, BERT is better at handling word polysemy by capturing word semantics in context.

\section{RESULTS}

\begin{table}[t]
\centering
\resizebox{\linewidth}{!}{
\begin{tabular}{lrrr}
\toprule
                 & \textbf{Prec.} & \textbf{Rec.} & \textbf{F1}   \\ \midrule
\textbf{CDR + PubMed Word}       & .8962 & .8797 & .8615 \\
\textbf{CDR + PubMed Flair}      & .9090 & .8984 & .8920 \\
\textbf{CDR + BioBERT} & .9030 & .8913 & .8971 \\ \cmidrule(lr){2-4}
\textbf{CDR + General Word} & .8046 & .8006 & .8026  \\
\textbf{CDR + General Flair} & .8794 & .8580 & .8686   \\
\textbf{CDR + BERT}       & \textbf{.9181} & \textbf{.9174} & \textbf{.9100} \\ \midrule
\textbf{CHEMDNER + PubMed Word}  & .8963 & .8887 & .8846 \\
\textbf{CHEMDNER + PubMed Flair} & \textbf{.9133 }& \textbf{.9112} & \textbf{.9018} \\
\textbf{CHEMDNER + BioBERT} & .9112 & .8861 & .8985  \\ \cmidrule(lr){2-4}
\textbf{CHEMDNER + General Word} & .8267 & .7570 & .7903 \\
\textbf{CHEMDNER + General Flair} & .8985 & .8696 & .8838  \\
\textbf{CHEMDNER + BERT}  & .9122 & .8840 & .8938 \\ \midrule
\textbf{CHEBI + PubMed Word}  & .7384 & .7123 & .7251 \\
\textbf{CHEBI + PubMed Flair}  & \textbf{.8051} & .7384 & .7703 \\
\textbf{CHEBI + BioBERT}  & .7858 & \textbf{.7703}  & \textbf{.7780 } \\ \cmidrule(lr){2-4}
\textbf{CHEBI + General Word} & .5999 & .6793	& .6372  \\
\textbf{CHEBI + General Flair}  & .7454	& .7196	& .7322  \\
\textbf{CHEBI + BERT}   & .7740	 & .7700 & .7720\\ \bottomrule
\end{tabular}}
\caption{CDR, CHEMDNER, and CHEBI Results.\vspace{-1em}}
\label{tab:source_max}
\end{table}

\begin{table*}[t]
\centering
\resizebox{.9\textwidth}{!}{
\begin{tabular}{@{}lrrrrrrrrr}
\toprule
                 & \multicolumn{3}{c}{\textbf{Male}}         & \multicolumn{3}{c}{\textbf{Female}}        &\multicolumn{3}{c}{\textbf{Difference}}        \\ \cmidrule(lr){2-4}  \cmidrule(lr){5-7}   \cmidrule(lr){8-10}  
\textbf{Dataset + Embeddings} & \textbf{Prec.} & \textbf{Rec.} & \textbf{F1}    & \textbf{Prec.} & \textbf{Rec.} & \textbf{F1}    & \textbf{Prec.} & \textbf{Rec.} & \textbf{F1}    \\ \midrule
\textbf{CDR + PubMed Word}       & 1         & .8230  & .9029 & 1         & .8230  & .9029 &  .0000         & .0000    & .0000 \\
\textbf{CDR + PubMed Flair}      & .9711     & .9486  & .9597 & .9344     & .9494  & .9418 & \textbf{.0367}    & -.0008 & \textbf{.0179}  \\ 
\textbf{CDR + BioBERT} & .8446 & .9044 & .8733 & .7764 & .9036 & .8352 & \textbf{.0682} & .0007  & \textbf{.0381} \\ \cmidrule(lr){2-10}
\textbf{CDR + General Word} & .9536 & .6756 & .7907 & .8530 & .6756 & .7539 & \textbf{.1006} & .0000	& \textbf{.0368} \\ 
\textbf{CDR + General Flair} & .8325 & .9400 & .8827 & .7610 & .9397 & .8408 & \textbf{.071}5 & .0003 & \textbf{.0419} \\
\textbf{CDR + BERT}       & .9867     & .8493  & .9128 & .9728     & .8444  & .9041 & \textbf{.0138}     & .0048  & .0087 \\ \midrule
\textbf{CHEMDNER + PubMed Word}  & .9990     & .8625  & .9257  & .9968     & .8622  & .9246  &  .0021     & .0003 &  .0011  \\
\textbf{CHEMDNER + PubMed Flair} & .9982     & .8836  & .9374 & .9885     & .8852  & .9340  & .0097     & -.007 & .0034 \\
\textbf{CHEMDNER + BioBERT} & .8847 & .8968 & .8907 & .8625 & .8963 & .8790 & \textbf{.0222} & .0005 & .0116 \\ \cmidrule(lr){2-10}
\textbf{CHEMDNER + General Word} & .9614 & .1966 & .3264 & .9311 & .1957 & .3233 & \textbf{.0302} & .0009 & .0030 \\
\textbf{CHEMDNER + General Flair} & .9559 & .8437 & .8963 & .9105 & .8433 & .8755 & \textbf{.0454} & .0004 & \textbf{.0208} \\
\textbf{CHEMDNER + BERT}  & .9913     & .8768  & .9306 & .9680     & .8762  & .9198 & \textbf{.0233}     & -.0006 & \textbf{.0107} \\  \midrule
\textbf{ASKDOC + PubMed Word}  & .9739       & .9330  & .9530  & .9739     & .9330  & .9530   & .0000     & .0000  & .0000   \\
\textbf{ASKDOC + PubMed Flair} & .8833     & .9523  & .9164 & .8278     & .9519  & .8852  & \textbf{.0555}    & .0005 & \textbf{.0312} \\
\textbf{ASKDOC + BioBERT} & .8026 & .9444 & .8677 & .7703	& .9443 & .8483 & \textbf{.0323} &	.0001 & \textbf{.0194} \\ \cmidrule(lr){2-10}
\textbf{ASKDOC + General Word} & .9681 & .6607 & .7854 & .9711 & .6604 & .7862 & -.0030 & .0003 & -.0008 \\
\textbf{ASKDOC + General Flair} & .8707 & .9491 & .9079 & .8166 & .9468 & .8765 & \textbf{.0542} & .0023 & \textbf{.0315} \\
\textbf{ASKDOC + BERT}  &  .9394         & .9288  & .9340 &  .8967    & .9282 & .9121 &  \textbf{.0427} & .0006 & \textbf{.0220} \\ \midrule
\textbf{CHEBI + PubMed Word}  & .9999	 & .8758	 & .9337  & .9979	 & .8715	 & .9305  & .0019	 & .0042	 & .0033 \\
\textbf{CHEBI + PubMed Flair}  & .9689	 & .9016	 & .9340  & .9545	 & .9031	 & .9281  &\textbf{ .0144}	& -.0015	 & .0060 \\
\textbf{CHEBI + PubMed BERT}  & .9170	 & .8673	 & .8914  & .8690	 & .8689	 & .8690  & \textbf{.0480} & -.0016	 & \textbf{.0225} \\ \cmidrule(lr){2-10}
\textbf{CHEBI + General Word}   & .9538	 & .5073	 & .6620  & .9147	 & .4956	 & .6424  & \textbf{.0391}	 & \textbf{.0118}	 & \textbf{.0196} \\
\textbf{CHEBI + General Flair}  & .9832	 & .8720	 & .9242  & .9677	 & .8701	 & .9163  & \textbf{.0155}	 & .0019	 & .0079 \\
\textbf{CHEBI + BERT}  & .9779	 & .8892	 & .9314  & .9223	 & .8882	 & .9048  & \textbf{.0556}	 & .0011	 & \textbf{.0266} \\
\midrule \midrule
\multicolumn{10}{c}{\textbf{Aggregate Measures}}  \\ \midrule
\textbf{AVERAGE PubMed/BioBERT} & .9370 & .8994 & .9155 & .9126 & .8994 & .9026 & \textbf{.0242} & \textbf{.0002} & \textbf{.0129} \\
\textbf{AVERAGE General} & .9479 & .7658 & .8238 & .9071 & .7637 & .8047 & .0407 & .0020 & .0191 \\ \midrule
\textbf{AVERAGE Word}   & .9763	 & .6919	 & .7850	 & .9548	 & .6897	 & .7771	 & \textbf{.0214}	 & .0022	 & \textbf{.0079} \\
\textbf{AVERAGE Flair}   & .9329	 & .9114	 & .9199	 & .8951	 & .9112	 & .8998	 & .0378	 & \textbf{-.0002}	 & .0201 \\
\textbf{AVERAGE (Bio)BERT}  & .9181	 & .8946	 & .9040	 & .8797	 & .8938	 & .8840	 & .0382	 & .0011	 & .0199\\
\bottomrule
\end{tabular}}
\caption{Synthetic (Template) Data Results. We \textbf{bold} the least biased aggregate measures and all differences greater than .01 to easily read the main findings. \vspace{-2em}}
\label{tab:synres}
\end{table*}

\paragraph{CDR, CHEMDNER, and CHEBI Results.}  Table~\ref{tab:source_max} reports the recall, precision, and F1 scores for each embedding type for the CDR, CHEMDNER, and CHEBI datasets. The reported scores are for the best models-hyperparameter combinations on their original validation datasets. Overall, we find that the Flair and BERT-based methods outperform word embeddings. The BERT embeddings result in the best performance for the CDR dataset. While in the CHEMDNER corpus, the PubMed Flair embeddings outperform the BERT embeddings (.9018 vs. .8938). For CHEMBI, the BioBERT embeddings work the best (.7720 vs. .7322 and .6372).

\paragraph{Synthetic (Template) Results.}
We evaluated several Named Entity Recognition (NER) models across multiple datasets and embeddings to assess gender bias, as summarized in Table~\ref{tab:synres}. Specifically, The aggregate measures in the bottom section of Table~\ref{tab:synres} highlight the overall trends in bias across embedding training data sources (PubMed vs. General) and embedding types (Word, Flair, and BERT). The bias analysis reveals that models generally perform differently on male versus female templates. Particularly,  PubMed-trained (including BioBERT) embeddings across all datasets show an average precision bias of .0242 against female-related names. The General embeddings exhibit substantially more bias, especially in precision with an average difference of .0407.  Moreover, while the average scores for Word and (Bio)BERT embeddings show less bias, the General and Flair embeddings indicate more significant bias in precision and F1 scores. These aggregate measures underscore the pervasive nature of gender bias in NER systems and the importance of addressing it in future work.

Overall, the major source of bias is that female-related names are being classified as chemicals. Intuitively, the word embeddings are less biased than Flair and (Bio)BERT-based embeddings because gender-related names are treated independently using word embeddings, or better, do not appear in the embeddings at all. This is particularly evident in the differences in performance between general word embeddings and the PubMed-based word embeddings. The PubMed embeddings do not generally have any direct mentions of named (e.g., John or Jane), hence they are generally less biased than the general domain.
 
 This finding that female-related names are classified as chemicals is consistent with prior research on naming conventions for brands being gendered~\cite{lieven2015effect}. To further investigate this, we randomly sampled 100 chemicals from all three datasets and measured the number of brand name mentions. Overall, we found one brand name in the CHEMDNER dataset, 19 in the CDR dataset, and 32 in the ASKDOC dataset, which generally matches the bias performance differences in Table~\ref{tab:synres} (i.e., biases are \textit{generally worse} in CDR and ASKDOC datasets than the CHEMDNER dataset).

\begin{table*}[t] 
\centering
\resizebox{.9\textwidth}{!}{
\begin{tabular}{lrrrrrrrrr}
\toprule
                 & \multicolumn{3}{c}{\textbf{Male}}         & \multicolumn{3}{c}{\textbf{Female}}        &\multicolumn{3}{c}{\textbf{Difference}}        \\ \cmidrule(lr){2-4}  \cmidrule(lr){5-7}   \cmidrule(lr){8-10}  
\textbf{Dataset + Embeddings} & \textbf{Prec.} & \textbf{Rec.} & \textbf{F1}    & \textbf{Prec.} & \textbf{Rec.} & \textbf{F1}    & \textbf{Prec.} & \textbf{Rec.} & \textbf{F1}    \\ \midrule
\textbf{CDR + PubMed Word}        & .8375     & .6023  & .7007 & .8206     & .6249  & .7095 & \textbf{-.0169}      & \textbf{.0226}   & .0088  \\
\textbf{CDR + PubMed Flair}       & .8614     & .6160  & .7183 & .8778     & .6702  & .7601 & \textbf{.0164}       & \textbf{.0542}   & \textbf{.0418}  \\
\textbf{CDR + BioBERT}            & .8303     & .6352  & .7198 & .8042     & .6693  & .7306 & \textbf{-.0261}      & \textbf{.0341}   & \textbf{.0108}  \\ \cmidrule(lr){2-10}
\textbf{CDR + General Word }      & .7538     & .6724  & .7108 & .7489     & .6986  & .7229 & -.0049      & \textbf{.0262}   & \textbf{.0121 } \\
\textbf{CDR + General Flair}      & .8479     & .6501  & .7359 & .8542     & .6707  & .7514 & .0063       & \textbf{.0206}   & \textbf{.0155}  \\
\textbf{CDR + BERT}               & .8742     & .6453  & .7425 & .8638     & .6589  & .7475 & \textbf{-.0104}      & \textbf{.0136}   & .0050  \\ \midrule
\textbf{CHEMDNER + PubMed Word}   & .8057     & .5966  & .6855 & .8158     & .6049  & .6947 & \textbf{.0101}       & .0083   & .0092  \\
\textbf{CHEMDNER + PubMed Flair}  & .8891     & .6155  & .7274 & .8871     & .6282  & .7356 & -.0020      & \textbf{.0127}   & .0082  \\
\textbf{CHEMDNER + BioBERT}       & .8537     & .6238  & .7208 & .8735     & .6434  & .7410 & \textbf{.0198}       & \textbf{.0196}   & \textbf{.0202}  \\ \cmidrule(lr){2-10}
\textbf{CHEMDNER + General Word}  & .7490     & .5546  & .6373 & .7975     & .5842  & .6743 & \textbf{.0485}       & \textbf{.0296}   & \textbf{.0370}  \\
\textbf{CHEMDNER + General Flair} & .8159     & .5678  & .6696 & .8821     & .6021  & .7157 & \textbf{.0662}       & \textbf{.0343}   & \textbf{.0461}  \\
\textbf{CHEMDNER + BERT}          & .7165     & .6315  & .6713 & .8309     & .6349  & .7198 & \textbf{.1144}       & .0034   & \textbf{.0485}  \\ \midrule
\textbf{CHEBI + PubMed Word}      & .7574     & .5998  & .6694 & .7548     & .6287  & .6860 & -.0026      & \textbf{.0289}   & \textbf{.0166}  \\
\textbf{CHEBI + PubMed Flair}     & .7540     & .6415  & .6932 & .7571     & .6740  & .7131 & .0031       & \textbf{.0325}   & \textbf{.0199}  \\
\textbf{CHEBI + BioBERT}          & .6896     & .5969  & .6399 & .7380     & .6148  & .6708 & \textbf{.0484}       & \textbf{.0179}   & \textbf{.0309}  \\ \cmidrule(lr){2-10}
\textbf{CHEBI + General Word}     & .6047     & .6541  & .6284 & .6132     & .6687  & .6397 & .0085       & \textbf{.0146}   & \textbf{.0113}  \\
\textbf{CHEBI + General Flair}   & .6066     & .5775  & .5917 & .6103     & .6001  & .6052 & .0037       & \textbf{.0226}   & \textbf{.0135}  \\
\textbf{CHEBI + BERT}             & .6274     & .6478  & .6374 & .6923     & .6467  & .6687 & \textbf{.0649}       & -.0011  & \textbf{.0313}  \\ \midrule \midrule
\multicolumn{10}{c}{\textbf{Aggregate Measures}}  \\ \midrule
\textbf{AVERAGE PubMed/BioBERT}   & .8087     & .6142  & .6972 & .8143     & .6398  & .7157 & \textbf{.0056}       & .0256   & \textbf{.0185}  \\ 
\textbf{AVERAGE General}          & .7329     & .6223  & .6694 & .7659     & .6405  & .6939 & .0330       & \textbf{.0182}   & .0245  \\  \midrule
\textbf{AVERAGE Word}             & .7514     & .6133  & .6720 & .7585     & .6350  & .6879 & \textbf{.0071}       & .0217   & \textbf{.0158}  \\
\textbf{AVERAGE Flair}            & .7958     & .6114  & .6894 & .8114     & .6409  & .7135 & .0156       & .0295   & .0242  \\
\textbf{AVERAGE (Bio)BERT}        & .7653     & .6301  & .6886 & .8005     & .6447  & .7131 & .0352       & \textbf{.0146}   & .0245  \\ \bottomrule
\end{tabular}}
\caption{AskDoc Results. We \textbf{bold} the least biased aggregate measures and all differences greater than .01 to easily read the main findings.\vspace{-2em}} 
\label{tab:real-results}
\end{table*}

\paragraph{AskDoc Results.}
The AskDoc results, as shown in Table~\ref{tab:real-results}, highlight various biases in chemical NER systems on real-world data. This table presents results from models trained on CDR, CHEMDNER, and CHEBI datasets, using different embeddings such as  Word, Flair, and (Bio)BERT. Again, the embeddings are both trained on general and domain-specific corpora (e.g., PubMed).

For the fine-grained results, we note that bias and performance can vary depending on unique combinations of the dataset and embedding types. However, for the aggregate results, we have two major findings. First, we find that general domain embeddings are more biased when applied to the chemical NER task (e.g., .0056 vs. .0330 precision). This further emphasizes the results from the synthetic data study. Second, we find that word embeddings are generally less fair than Flair BERT/BioBERT embeddings for precision (.0071 vs. .0156 and .0352) and F1 (.0158 vs. .0242 and .0245). 

What does this mean in real-world terms? Considering a sample of 1,000,000 chemical mentions across male and female posts (a relatively small number in social media), a 4\% recall difference results in an additional 40,000 false negatives for the female group. For example, there are well-known health disparities between men and women for depression, with absolute differences of less than 3\%~\cite{salk2017gender}. Hence, a 4\% recall difference can substantially impact findings if applied researchers or practitioners use out-of-domain models to understand medications for this disease. Such a considerable gap can markedly affect the utility and trustworthiness of these predictive outcomes in practical scenarios.

\begin{figure}
    \centering
    \includegraphics[width=\linewidth]{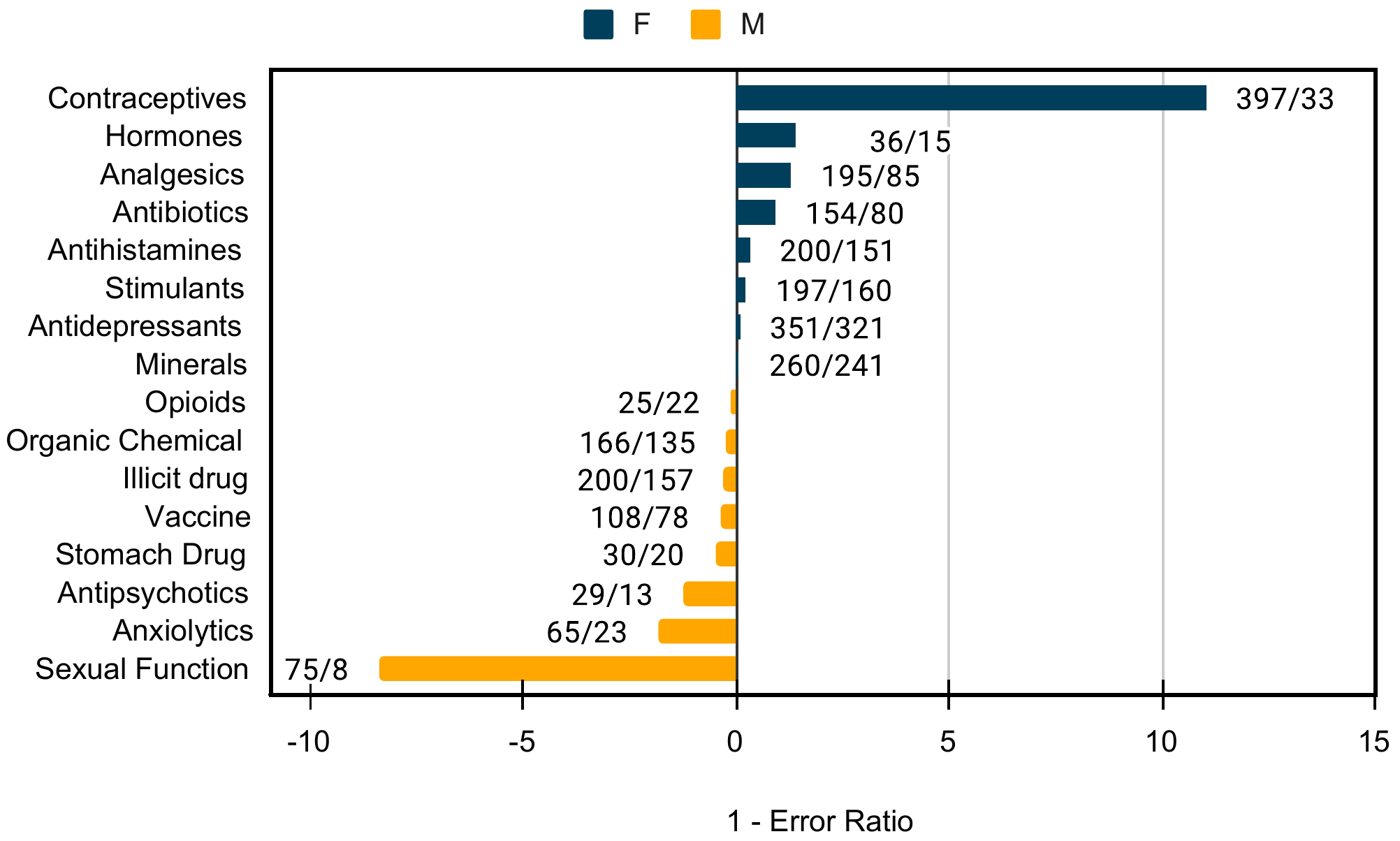}
    \caption{Ratio of false negatives for various drug categories. The ratio is represented next to each bar. For female-leaning errors, the female false negative count ($FN^k_f$) is in the numerator. For male-leaning errors, the male false negative count ($FN^k_m$) is in the numerator.\vspace{-2em}}
    \label{fig:errors}
\end{figure}

\paragraph{AskDoc Error Analysis.}
Our experiments show that Chemical NER systems are biased. However, what specifically is causing the errors? For the synthetic data, the answer is gender-related names. To understand the errors in the AskDoc data, we analyzed the errors made by the best NER models trained on the out-of-domain corpus (CHEMDNER and CDR) and tested the male and female splits of the AskDocs corpus. In Figure~\ref{fig:errors}, we report the ratio of false negatives for different categories of drugs/chemicals. For every false negative made by the top models of each dataset-model combination, we manually categorized them into a general chemical class (e.g., Contraceptives, Analgesics/Pain Killers, and Stimulants). Formally, let $FN^k_m$ represent the total number of false negatives for chemical types $k$ and male data $m$. Let $FN^k_f$ represent the female false negatives. If $FN^k_m$ is larger than $FN^k_f$, we define the ratio as $-(1 - FN^k_m / FN^k_f)$. Likewise, if $FN^k_f$ is greater than $FN_m$, then we define the ratio as $1-(FN^k_f / FN^k_m)$. Hence, when male ratios are higher, the score is negative; otherwise, it is positive.

Overall, we make several important findings. First, we find that the models make slightly more false negatives on the chemicals categories Contraceptives (e.g., birth control and Plan B One-Step), Hormones (e.g., Megace used to treat the symptoms of loss of appetite and wasting syndrome in people with illnesses such as breast cancer), Analgesics (i.e., Pain Killers such as Tylenol) and Antibiotics on the female dataset. In contrast, the models make slightly more errors in the chemical categories Anxiolytics (e.g., drugs used to treat anxiety), Antipsychotics (e.g.,  chemicals used to manage psychosis, principally in schizophrenia), and sexual function drugs (e.g., Viagra). Furthermore, while the ratio for the most male- and female-related errors (Contraceptives and Sexual Function) are similar, the absolute magnitudes are substantially different. For instance, there are 397 Contraceptive $FN$s in the female dataset, but only 75 Sexual Function $FN$s appear in the male dataset. This provides an explanation for the large differences in recall on the AskDoc corpus between the male and female datasets.



\section{CONCLUSION}

In this paper, we evaluate the gender bias of Chemical NER systems. Moreover, we compare bias measurements from synthetic data with real-world self-identified data. We make two major findings. First, Chemical NER systems are biased with regard to gender for synthetic data. Specifically, our study found that \textbf{female name-like patterns feature prominently in chemical naming conventions}. This characteristic leads to a notable bias in NER systems, where female-related names are disproportionately identified as chemicals, inadvertently escalating the gender bias in these systems. Second, we explored the performance of these models in real-world scenarios and found that most models perform better on male-related data than female-related data. \textbf{A striking revelation was the system's poor performance when identifying chemicals frequently found in female-related data, such as mentions of contraceptives.}

In conclusion, the results of our study emphasize the urgent need for deliberate bias mitigation strategies in Chemical NER systems. Our findings spotlight the necessity for incorporating both synthetic and real-world data considerations to develop models that are both fair and reliable. There are two major paths for future research. First, while large language models are still behind in terms of performance for NER systems~\cite{wang2023gpt}, they are becoming more common. Future work should explore biases in prompting-based NER solutions. Second, we plan to explore how the chemical NER biases impact downstream tasks such as relationship classification and question answering.

\section*{ACKNOWLEDGEMENTS}
 This material is based upon work supported by the National Science Foundation (NSF) under Grant~No. 1947697 and 2145357.

\bibliography{anthology}
\bibliographystyle{acl_natbib}

\appendix

\section{Appendix}
\label{sec:appendix}

\subsection{Datasets}
\paragraph{CDR~\cite{li2016biocreative}} We use the BioCreative V CDR shared task corpus. The CDR corpus comprises 1,500 PubMed articles with 4,409 annotated chemicals, 5,818 diseases, and 3,116 chemical disease interactions. This corpus is designed to address two distinct tasks: Relation classification and NER. For this study, we focus on the NER for chemical entities. The annotator agreement for this corpus was .87. Finally, we used the same train, validation, and test splits from the shared task for our experiments.

\paragraph{CHEMDNER~\cite{krallinger2015chemdner}} The CHEMDNER corpus includes abstracts from 10000 chemistry-related journals published in 2013 on PubMed. Each abstract was manually annotated for chemical mentions. These mentions were categorized into seven subtypes: abbreviation, family, formula, identifier, multiple, systematic, and trial. The BioCreative organizers divided the corpus into training (3500 abstracts), development (3500 abstracts), and test (3000 abstracts) sets. The BioCreative IV CHEMDNER corpus comprises 84,355 chemical mention annotations across 10,000 abstracts, with an inter-annotator agreement of .91~\cite{krallinger2015chemdner}. For this study, we only use the major Chemical annotations and ignore the subtypes for consistency across corpora. Finally, we use the same train, validation, and test splits used in the shared task for our experiments.

\paragraph{CHEBI~\cite{shardlow2018new}.} 
We also use the ChEBI corpus, an extensive dataset consisting of 199 annotated abstracts and 100 full papers. This corpus contains over 15,000 named entity annotations and more than 6,000 inter-entity relations, specifically aligned with the needs of the ChEBI database curators. The dataset has annotated chemicals, proteins, species, biological activities, and spectral data. Moreover, it has a high inter-annotator agreement of 0.80-0.89 (F1 score, strict-matching). It also categorizes relationships into several types such as Isolated From, Associated With, Binds With, and Metabolite Of, offering a detailed view of the interactions between metabolites and other entities. This corpus is not only a rich source for exploring lexical characteristics of metabolites and associated entities but also serves as a critical resource for training machine learning algorithms in the recognition of these entities and their relations in the biochemical context.

\subsection{Hyper-Parameter Settings}
In this section, we report the best hyperparameter for each model. Similar to random hyperparameter search~\cite{bergstra2012random}, we generate 100 samples using different parameters for each dataset-model combination (e.g., we generate 100 versions of BERT for the CDR dataset). For the specific hyper-parameters, we used sample dropout from .1 to .9, hidden layer sizes from $\{$128, 256, 512, 1024$\}$, learning rates selected from 1e-4 to 1e-1 at random, and the option of whether to fine-tune the embedding layers (i.e., True vs. False). In addition, we trained all models for 25 epochs with a mini-batch size set to 32, where only the best model on the validation dataset is saved after each epoch. Finally, all experiments were run on four NVidia GeForce GTX 1080 Ti GPUs.

\subsection{Error Analysis and Discussion} 

\begin{table*}[t]
\centering
\begin{tabular}{lrrrrr}
\toprule
 &
 \textbf{Total Male} &
  \textbf{FNR Male} &
  \textbf{Total Female} &
  \textbf{FNR Female}  \\ \midrule
\textbf{Contraceptives}       & 33  & 1.0000 & 408 & .9730   \\
\textbf{Hormones}             & 170  & .0882  & 230 & .1565  \\
\textbf{Analgesics}           & 571  & .1489  & 952 & .2048  \\
\textbf{Antibiotics}          & 326  & .2454  & 347 & .4438 \\
\textbf{Antihistamines}       & 270  & .5593  & 295 & .6780 \\
\textbf{Stimulants}           & 522 & .3065  & 390 & .5051  \\
\textbf{Antidepressants}      & 781 & .4110  & 1043 & .3365 \\
\textbf{Minerals}             & 605  & .3983  & 785 & .3312  \\
\textbf{Opioids}              & 43  & .5814  & 95 & .2316 \\
\textbf{Organic Chemical}     & 441 & .3764  & 346 & .3902 \\
\textbf{Illicit drug}         & 353 & .5666  & 311 & .5048   \\
\textbf{Vaccine}              & 108  & 1.0000 & 78 & 1.0000 \\
\textbf{Stomach Drug}         & 55   & .5455  & 44 & .4545  \\
\textbf{Antipsychotics}       & 47  & .6170  & 95 & .1368   \\
\textbf{Anxiolytics}          & 126  & .5603  & 100 & .2300  \\
\textbf{Sexual Function Drug} & 78  & .9615  &  8 & 1.0000 \\ \cmidrule(lr){2-3} \cmidrule(lr){4-5}
\textbf{PCC between Total and FNR} &  \multicolumn{2}{c}{\textbf{-.58}} &  \multicolumn{2}{c}{-.26} \\ \bottomrule
\end{tabular}
\caption{False negatives rate (FNR) for female and male-related AskDoc datasets. The pearson correlation coefficient (PCC) between the frequency of each chemical type and the FNR for teach group is marked in the last row.\vspace{0em}}
\label{tab:fnrs}
\end{table*}



Interestingly, we find that the prevalence of chemicals across gender-related posts matches the prevalence found in traditional biomedical studies. Previous research report that women have been prescribed analgesics (e.g., pain killers such as opioids) twice as often as  men~\cite{chilet2014gender,serdarevic2017gender}. While there is still limited understanding about whether men are under-prescribed or women are over-prescribed, the disparities in prescriptions are evident. Thus, the finding in Figure~\ref{fig:errors} that we receive twice as many analgesics $FN$s for female data is important. Depending on the downstream application of the Chemical NER system, these performance disparities may potentially increase harm to women. For example, if more varieties of drugs are prescribed to women, but our system does not detect them, then an ADR detection system will not be able to detect important harms. 


We also find differences in Antibiotic $FN$s in Figure~\ref{fig:errors}. There have also been medical studies showing gender differences in Antibiotic prescriptions. For example, a recent meta-analysis of primary care found that women received more antibiotics than men, especially women aged 16–54, receiving 36\%–40\% more than males of the same age~\cite{smith2018understanding}. Again, if we do not detect many of the antibiotics prescribed to women, this can cause potential health disparities in downstream ADR (and other) systems.

Next, in Table~\ref{tab:fnrs}, we report the false negative rate (FNR) for each category along with the general frequency of each category. Using the Pearson correlation coefficient, we relate the frequency of each category with the false negative rate for the male and female groups, respectively. Intuitively, we would expect the false negative rate to go down as the frequency increases, which matches our findings. However, we find that the correlation is much stronger for the male group than the female group.

In Table~\ref{tab:wfnr}, we report the FNR for the female and male groups, respectively. We also introduce a new metric, weighted FNR, which assigns importance scores for each of the FNRs shown to create a macro-averaged metric.  Intuitively, the distribution of categories is different for both the male and female groups. So, we want to test whether the FNR scores are distributed uniformly across all categories, irrespective of, whether the errors are more concentrated for gender-specific categories. More errors in gender-specific categories can adversely impact a group that is not captured with the global FNR metric. Formally, we define wFNR for the female group as
\begin{equation*}
    wFNR^f = \sum_i^N w^f_i FNR^f_i
\end{equation*}
where $FNR^f_i$ represents the female false negative rate for category $i$. Likewise, $w^f_i$ is defined as
\begin{equation*}
    w^f_i = \frac{1}{\sum_i w_i^f} \cdot \frac{N^f_i / N^f}{N^m_i/N^m}
\end{equation*}
where $N^f_i$ and $N^m_f$ represent the total number of times a category $i$ appears for the female and male groups, respectively. Intuitively, we are dividing the ratio of each category for female and male groups. So, if a category appears more often for females than males, proportionally, then the score will be higher. We normalize these scores for each group so they sum to one. Overall, we find an absolute gap of more than 1\% (3\% relative difference) between the FNR for male and female groups. But, even worse, there is a much larger gap (.1213 vs .0116) when using wFNR. This result suggests that many of the false negatives are concentrated for gender-specific categories (e.g., contraceptives) for the female group more than the male group.


\begin{table}[t]
\centering
\resizebox{.21\textwidth}{!}{%
\begin{tabular}{lrr}
\toprule
       & \textbf{FNR}    & \textbf{wFNR}   \\ \midrule
\textbf{Male}   & .3948  & .6875  \\
\textbf{Female} & \textbf{.4064}  & \textbf{.8088}  \\ \midrule
\textbf{Gap}    & .0116  & .1213  \\
\textbf{Ratio}  & 1.0294 & 1.1764 \\ \bottomrule
\end{tabular}%
}
\caption{FNR and weighted FNR (wFNR) results.}
\label{tab:wfnr}
\end{table}





\subsection{Limitation}

There were several limitations to our study. First, the adjudication of disagreeing items was dependent on the judgment of a single graduate student, potentially introducing human error and bias compared to a multi-adjudicator approach. Second, the vast volume of data from the active r/AskDoc subreddit community makes the feasibility of one person's comprehensive review debatable. Although our annotation method is in line with standard practices, a more multi-faceted approach involving numerous annotators and adjudicators might offer improved accuracy and consistency in future datasets. Third, our study focuses on binary representations of gender (ignoring non-binary people). Moreover, the SSN names may not adequately mention immigrant-related names. Hence, the results may be European-specific.
\end{document}